\documentclass[10pt,twocolumn,letterpaper]{article}

\usepackage{wacv}
\usepackage{times}
\usepackage{epsfig}
\usepackage{graphicx}
\usepackage{amsmath}
\usepackage{amssymb}

\usepackage{subfig}
\usepackage{booktabs}
\usepackage{amsmath}

\usepackage{pifont}
\newcommand{\cmark}{\ding{51}}%
\wacvfinalcopy 

\ifwacvfinal
\def\assignedStartPage{1} 
\fi

\ifwacvfinal
\usepackage[breaklinks=true,bookmarks=false]{hyperref}
\else
\usepackage[pagebackref=true,breaklinks=true,colorlinks,bookmarks=false]{hyperref}
\fi

\ifwacvfinal
\else
\fi

\begin{document}

\title{Graph Relation Transformer: Incorporating pairwise object features into the Transformer architecture}

\author{Michael Yang\\
{\tt\small myang2@alumni.cmu.edu}
\and
Aditya Anantharaman\\
{\tt\small adityaan@alumni.cmu.edu}
\and
Zachary Kitowski\\
{\tt\small zkitowsk@alumni.cmu.edu}
\and
Derik Clive Robert\\
{\tt\small dclive@alumni.cmu.edu}\\
\\
Carnegie Mellon University\\
Pittsburgh, USA\\
}

\maketitle

\begin{abstract}
    Previous studies such as VizWiz find that Visual Question Answering (VQA) systems that can read and reason about text in images are useful in application areas such as assisting visually-impaired people. TextVQA is a VQA dataset geared towards this problem, where the questions require answering systems to read and reason about visual objects and text objects in images. One key challenge in TextVQA is the design of a system that effectively reasons not only about visual and text objects individually, but also about the spatial relationships between these objects. This motivates the use of `edge features', that is, information about the relationship between each pair of objects. Some current TextVQA models address this problem but either only use categories of relations (rather than edge feature vectors) or do not use edge features within the Transformer architectures. In order to overcome these shortcomings, we propose a Graph Relation Transformer (GRT), which uses edge information in addition to node information for graph attention computation in the Transformer. We find that, without using any other optimizations, the proposed GRT method outperforms the accuracy of the M4C baseline model by 0.65\% on the val set and 0.57\% on the test set.
    Qualitatively, we observe that the GRT has superior spatial reasoning ability to M4C.\footnote{The code used to obtain our results can be found at \url{https://github.com/michaelzyang/graph-relation-m4c} and \url{https://github.com/derikclive/transformers}}
\end{abstract}

\section{Introduction}

\begin{figure}
    \centering
    \includegraphics[scale=0.6]{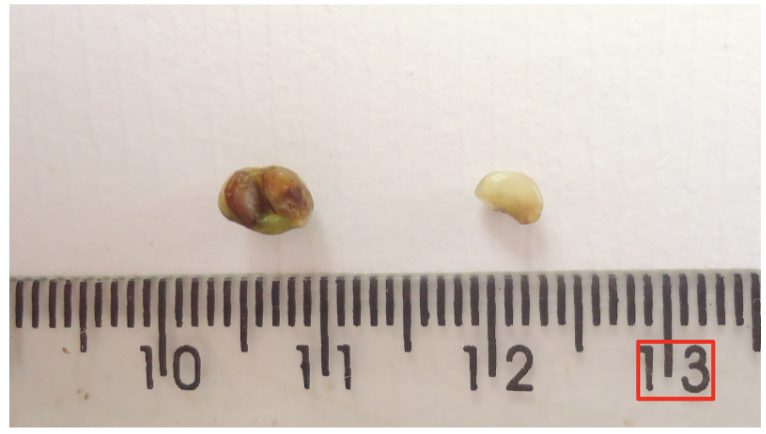}
    \caption{An example from the TextVQA dataset which shows the importance of reasoning about spatial relationships between objects.\textbf{Q} What is the last number to the right? \textbf{Ans} 13.}
    \label{fig:example_textvqa}
\end{figure}

Visual Question Answering (VQA) is the task of answering questions by reasoning over the question and the image corresponding to the question. Although VQA models have shown a lot of improvement in recent years, these models still struggle to answer questions which require reading and reasoning about the text in the image. This is an important problem to solve because studies have shown that visually-impaired people frequently ask questions which involve reading and reasoning about the text in images. The `TextVQA' task \cite{lorra} is a VQA task where the questions are focused on the text in images as shown in Figure \ref{fig:example_textvqa}. In order to answer this question, the model needs to understand that the visual object `13' is the last object among all objects towards the right of the image. This leads to the introduction of an additional modality involving the text in the images which is often recognized through Optical Character Recognition (OCR). The incorporation of this additional modality enhances the difficulty of this task as compared to a standard VQA task. 


The current state-of-the-art models \cite{gao2020structured, kant2020spatially} in this domain use graph attention paired with multimodal Transformer based approaches to model relationships between image objects, OCR objects and question tokens. Both approaches try to capture the relationship between these objects within the image using rich edge features. However, we see that they either use edge features for representation learning before the Transformer layers \cite{gao2020structured} or do not incorporate rich edge features when graph attention computation is done in Transformer layers \cite{kant2020spatially}. These rich edge features can be leveraged within the Transformer layers to allow for better representations between image and OCR objects. 

In this work, we propose a novel Graph Relation Transformer (GRT) which uses rich, vector-based edge features in addition to node information for graph attention computation in the Transformer. The proposed GRT outperforms the M4C baseline model \cite{m4c} while also improving the spatial reasoning ability of the model. We also provide qualitative examples of cases where our proposed approaches performs better than the M4C baseline model \cite{m4c}.


\section{Related Work}  
\label{related_work}

\subsection{VQA}
For VQA tasks, the main multimodal challenges are how to represent the visual and language modalities and how to fuse them in order to perform the Question Answering (QA) task. In terms of representing the questions, word embeddings such as GloVe \cite{pennington2014glove} are commonly used in conjunction with recurrent neural networks (RNNs) such as Long Short-Term Memory (LSTM) networks \cite{Gers01lstm}, for example by Fukui \etal~\cite{fukui16mcb}. For representing the visual modality, grid-based Convolutional Neural Networks (CNNs) such as Resnet \cite{he2015deep} are often used as visual feature extractors.

For representation, Bottom-up And Top-down Attention (BUTD) \cite{Anderson_2018_CVPR} is a canonical VQA method. Previous methods used a fixed-size feature map representation of the image, extracted by grid-based CNN, whereas BUTD uses a Region-CNN such as Faster R-CNN \cite{girshick2015fast} to propose a number of variably-sized objects, bottom-up, over which the question representation will attend, top-down.

Applying several tweaks to the BUTD model, the Pythia model \cite{jiang2018pythia} improved its performance further. Pythia reported performance gains from ensembling with diverse model setups and different object detectors, architecture, learning rate schedule and data augmentation and fine-tuning on other datasets.

Bilinear Attention Networks (BANs) \cite{KimBAN} aim to improve multimodal fusion by allowing for attention between every pairwise combination of words in the question and objects in the image. This setup is motivated by the common occurrence that different words in the question refer to different objects in the image and thus allowing for bilinear interaction improves grounding between the modalities. Low-rank matrix factorization is used to keep the computation cost tractable and the Multimodal Residual Network (MRN) \cite{KimBAN} method is used to combine all the representations from the bilinear attention maps. 

\subsection{TextVQA}
\paragraph{Fusion Methods} 
Most of the methods proposed for TextVQA involve the fusion of features from the different modalities - Text, OCR and Images. LoRRA \cite{lorra} and M4C \cite{m4c} are two such model architectures that primarily use attention mechanisms to capture the interactions between the inputs from different modalities. In addition, both of these models employ a pointer network \cite{vinyals2015pointer} module to directly copy tokens from the OCR inputs. This enables the models to reduce the number of out-of-vocabulary words predicted by supplementing the fixed answer vocabulary with the input OCR context.

LoRRA \cite{lorra} was proposed as the baseline for the 2019 TextVQA challenge and is composed of three major components - one to combine image and question features, another to combine OCR and question features and the third one to generate the answer. M4C improves the fusion of the input modalities through the use of Multimodal Transformers which allows both inter-modal and intra-modal interactions. The multimodal Transformer uses features from all three modalities and uses a pointer-augmented multi-step decoder to generate the answer one word at a time unlike the LoRRA model that uses a fixed answer vocabulary.

\paragraph{Graph Attention Methods}
\vspace{-1em}
Graph attention networks are another approach to try to capture the relationships between objects detected within an image. This approach was first used by \cite{regat} to solve the VQA task by detecting objects within an image and then treating each of these objects as nodes within a graph. This approach uses several iterations of modifying each node's representation with that of its neighbors'. This approach produced better results than existing methods at the time. Extending this approach to the TextVQA task, detected OCR tokens were added as additional nodes within these graphs.

The Stuctured Multimodal Attention (SMA) approach \cite{gao2020structured} won the 2020 CVPR TextVQA challenge using a variation of the standard graph attention described above. They used various attention mechanisms to create one attended visual object representation and one attended OCR object representation, which were then fed into the Multimodal Transformer layers instead of the individual objects. Importantly, these attention mechanisms involve the use of edge features between objects, an idea which we implement differently in our GRT.  
A limitation of this work is that using edge features for representation learning before the Transformer layers does not leverage the graph computation inherent to Transformers. It also does not create contextualized representations for each object unlike those that can be created using the Transformer layers.

The Spatially Aware Multimodal Transformers (SAMT) approach was the 2020 CVPR TextVQA runner-up \cite{kant2020spatially}. They created a spatial relation label for every pair of objects and restricted each Transformer attention head to attend over paired objects with certain labels only. A limitation of this work is that there is no natural way to make use of rich edge features since masking is based on discrete edge categories, not continuous / vector edge features. 

We propose to overcome the limitations of both SMA \cite{gao2020structured} and SAMT \cite{kant2020spatially} in this work using a novel Graph Relation Transformer which uses rich edge information for graph attention computation in the Transformer. Integrating rich edge information within the Transformer self-attention layer helps us overcome the limitation of SMA \cite{gao2020structured}. Continuous space, rich edge features helps us overcome the limitation of SAMT \cite{kant2020spatially}.

\subsection{Graph Networks outside VQA}

Graph networks have also been proposed for tasks outside the VQA domain. Cai et al. \cite{cai2020graph} propose a graph transformer for the graph to sequence learning task. They use a learned vector representation to model the relation between nodes. Yun et al. \cite{yun2019graph} propose graph transformer networks based on graph convolution networks for node classification tasks. Unlike our proposed approach, they do not use rich edge features between objects based on the appearance and position of objects in the image which provide valuable information in image-based tasks like the TextVQA task. There have also been efforts to generalize the Transformer architecture for arbitrary graphs related tasks. \cite{yun2019graph} propose modifications to the original transformer architecture by leveraging the inductive bias present in graph topologies while also injecting edge information into the architecture.

\section{Proposed Approach}
\label{proposed_approach}

\begin{figure*}
    \subfloat[Appearance Similarity edge feature]{{\includegraphics[height=6.5cm]{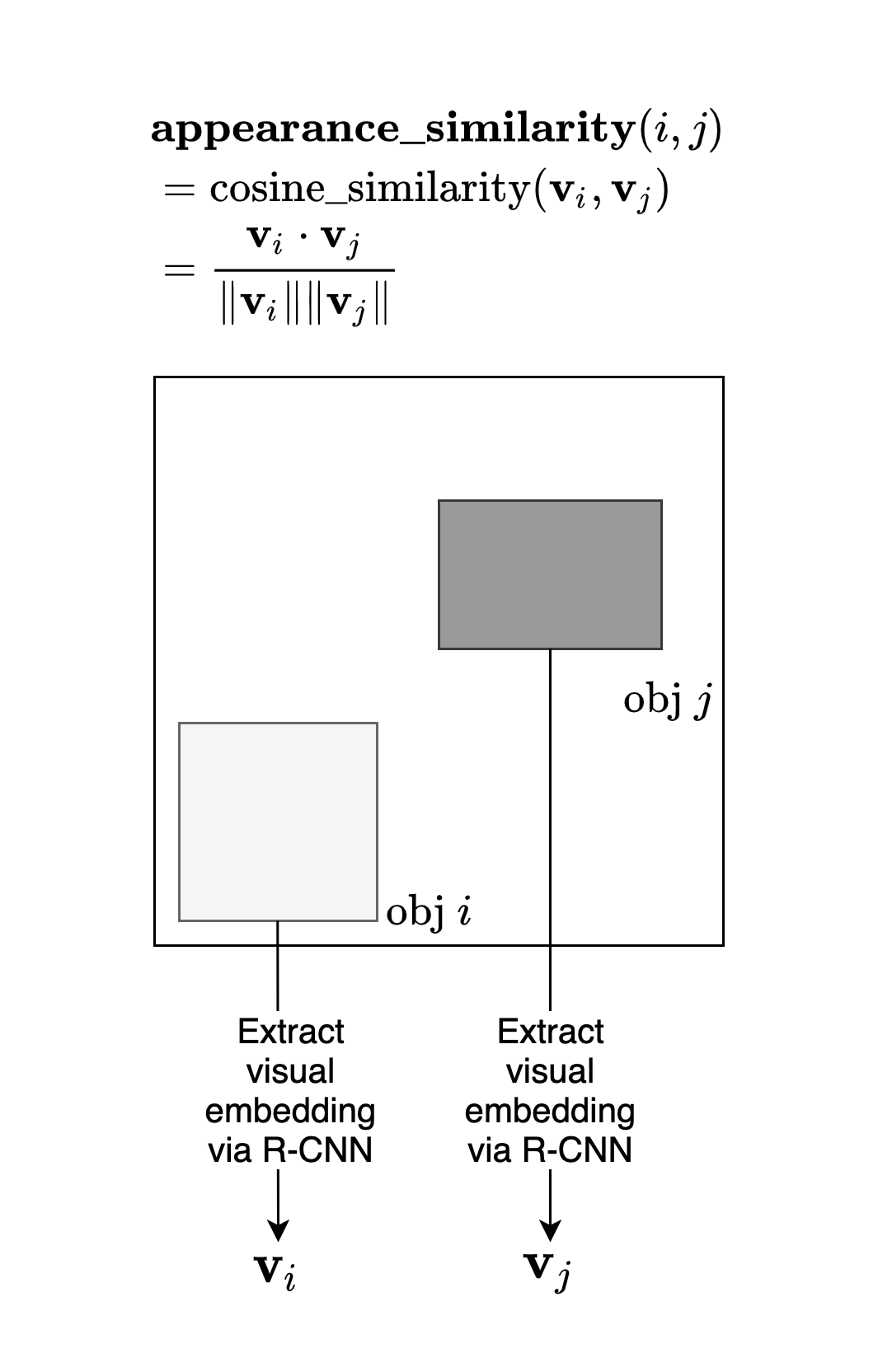}}}
    \hspace{2mm}
    \subfloat[Spatial Translation edge feature]{{\includegraphics[height=6.5cm]{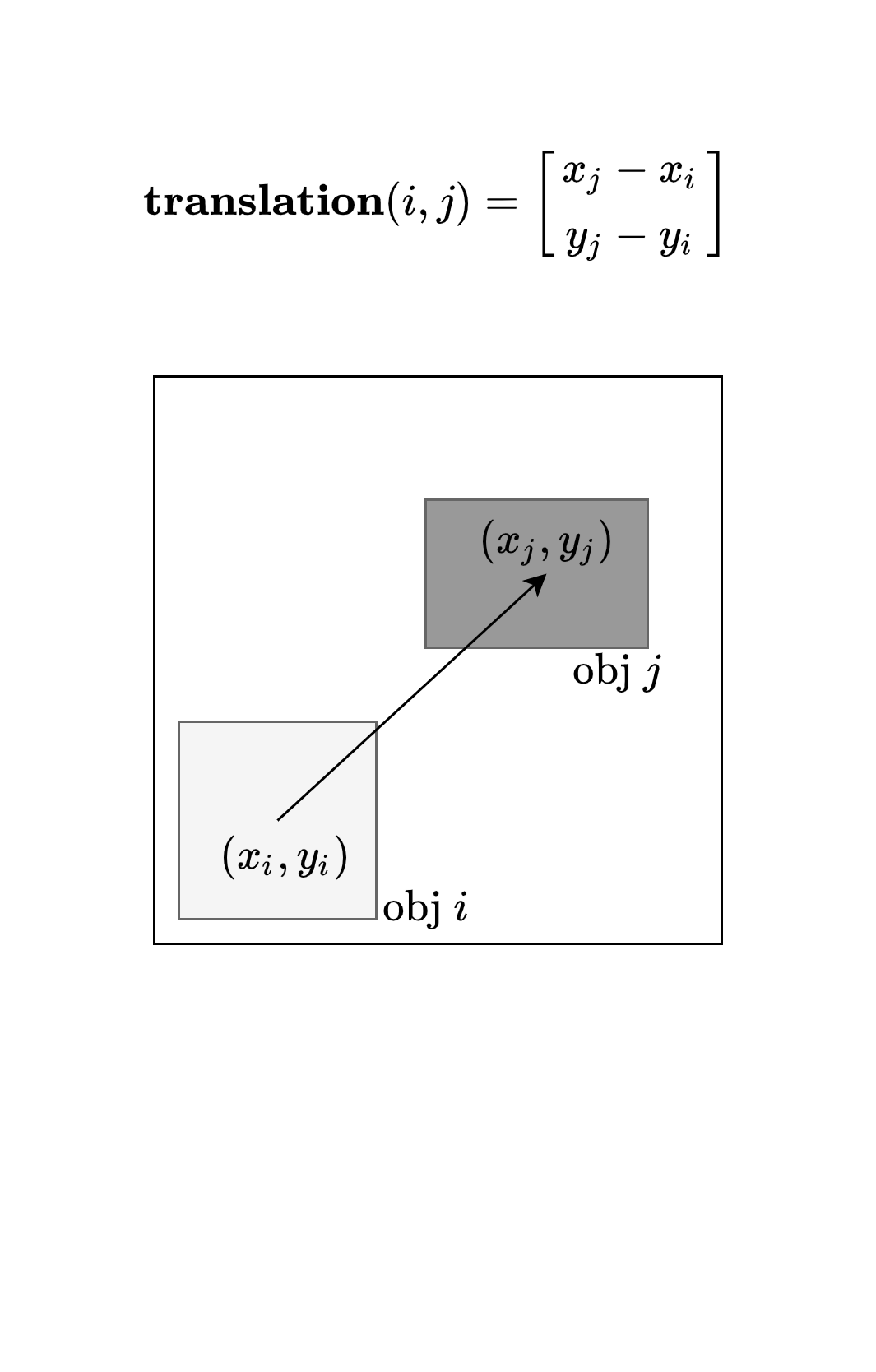}}}
    \hspace{2mm}
    \subfloat[Spatial Interaction Label edge feature]{{\includegraphics[height=6.5cm]{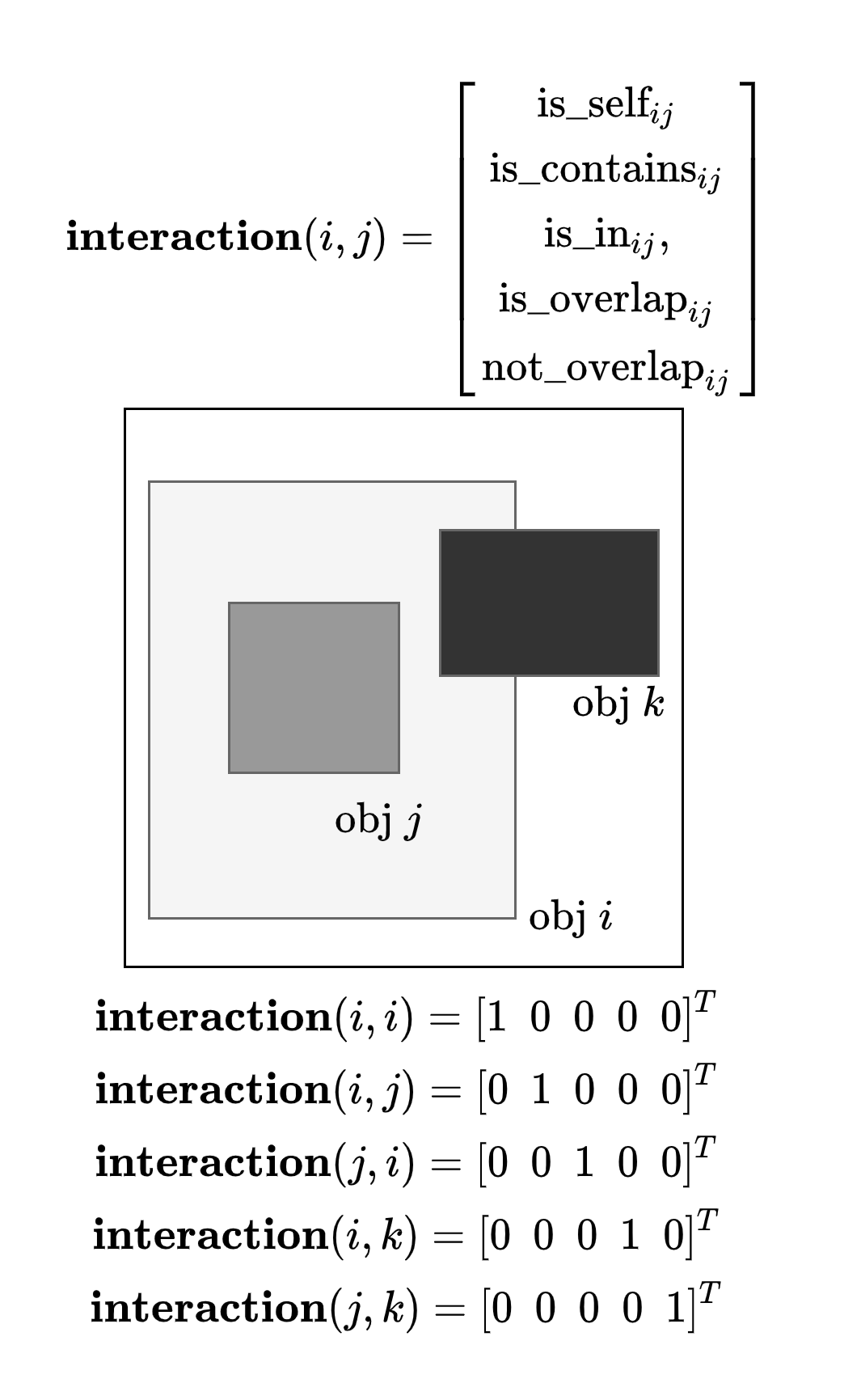}}}
    \hspace{2mm}
    \subfloat[Modality Pair Label edge feature]{{\includegraphics[height=6.5cm]{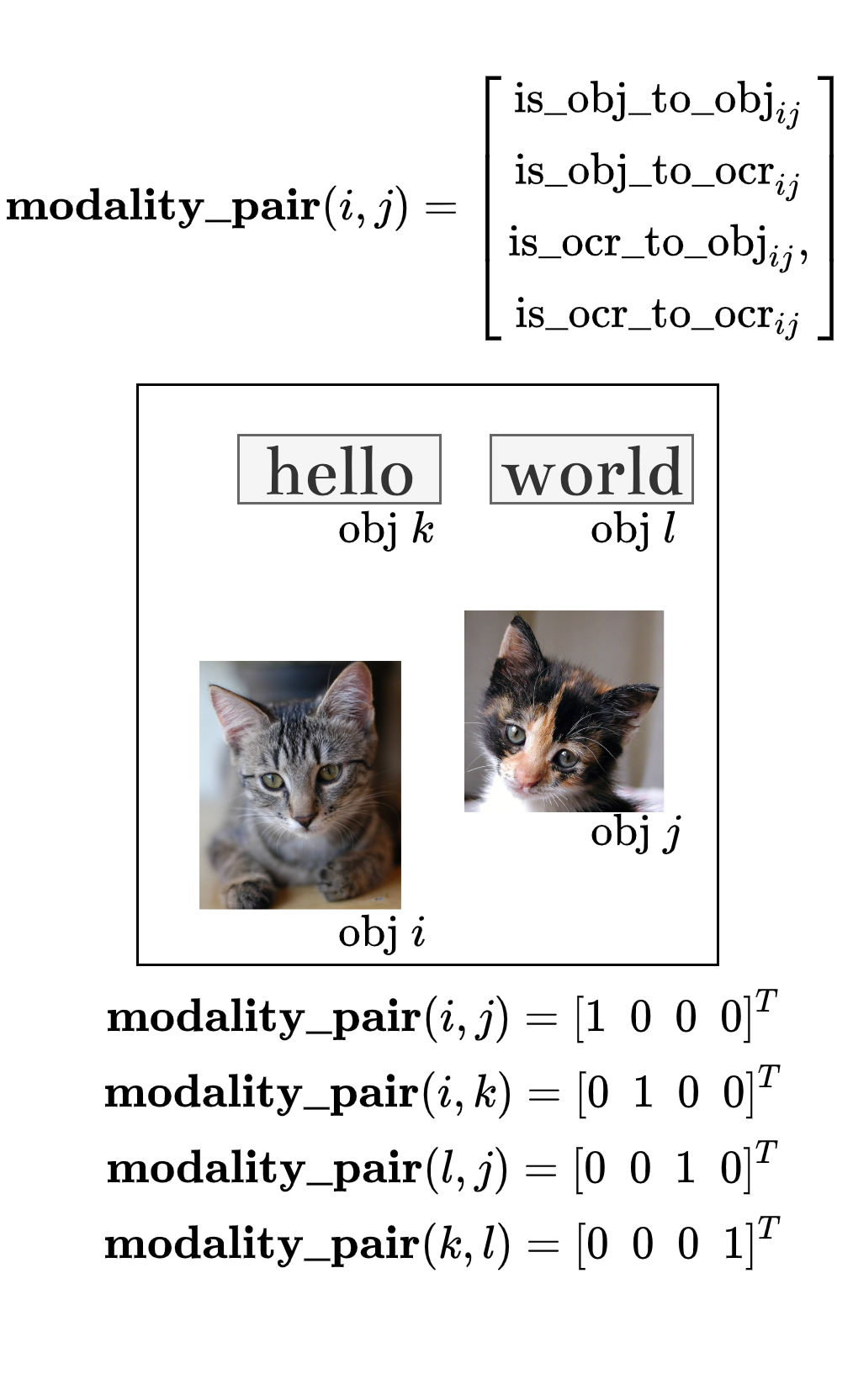}}}
    
    \caption{Illustration of the four edge features we propose to use for the TextVQA task. For each pair of visual objects $i$ and $j$ in an image, we extract a vector of edge features $\textbf{e}_{ij}$ by concatenating these four edge features: \textbf{appearance\_similarity}$(i,j)$, \textbf{translation}$(i,j)$, \textbf{interaction}$(i,j)$ and \textbf{modality\_pair}$(i,j)$.}\label{fig:edge_features}
\end{figure*}

In this section, we motivate and explain GRT, our novel extension to the M4C model. 


\subsection{Graph Relation Transformer (GRT)}

A common theme among the top performing models \cite{gao2020structured,kant2020spatially} is extracting and using graph relations between detected objects in images. However, no prior work in the VQA domain has used edge features within the Transformer layers for representation learning. Here, we will describe four types of edge features with which we will experiment and how they will be used in the Transformer. A key observation of the Transformer Encoder architecture is that the self-attention inherently performs graph computation across each input as if the inputs were structured as a fully-connected graph, as each input attends over all inputs. However, there is no natural place to inject edge relations in the Transformer Encoder layer. To achieve this, we propose Graph Relation Self-Attention in Transformers.

\vspace{-2 mm}
\paragraph{Edge features}
We hypothesize that four types of edge features will be useful for the model to better understand the relationships between objects. Each of these features are defined for a pair of objects (which may be from the visual modality or the OCR modality). These are (1) Appearance similarity (2) Spatial translation feature (3) Spatial interaction labels (4) Modality pair labels. Figure \ref{fig:edge_features} illustrates how these are extracted for every pair of visual objects in an image.

Appearance similarity may be especially useful for the OCR modality, as OCR tokens in an image that belong together in a sentence or paragraph usually have similar font and overall appearance. We use the cosine similarity between the 2,048-dimensional Faster R-CNN embeddings as the (1-dimensional) appearance similarity feature. This edge feature is illustrated in Figure \ref{fig:edge_features}a.

The spatial translation feature from object $i$ to object $j$ is simply a 2-dimensional feature that is the translation from the center of object $i$'s bounding box to object $j$'s in the $x$ and $y$ directions, where these translations are normalized to the length and width of the object such that this feature's range is [-1, 1] in both directions. Clearly, this feature provides the model information about the relation between objects in terms of spatial direction. This edge feature is illustrated in Figure \ref{fig:edge_features}b.

Similar to the labels proposed by Kant \etal~\cite{kant2020spatially}, we use 5 mutually exclusive class labels indicating different types of spatial interaction [\textit{is\_self}, \textit{is\_contains}, \textit{is\_in}, \textit{is\_overlap}, \textit{not\_overlap}], where \textit{is\_self} indicates a self-edge between object $i$ to itself, \textit{is\_contains} indicates if object $i$'s bounding box completely contains object $j$'s, \textit{is\_in} indicates if object $j$'s bounding box completely contains object $i$'s, \textit{is\_overlap} indicates if there is any overlap between the bounding boxes of objects $i$ and $j$ and they do not fall under any of the previous classes and \textit{not\_overlap} indicates if there is no overlap between the two objects' bounding boxes.\footnote{We do not include the 8 directional labels used by Kant \etal~\cite{kant2020spatially} as this information should be provided by the previously mentioned spatial translation feature.} These labels provide the model information about how objects interact with each other in space. This edge feature is illustrated in Figure \ref{fig:edge_features}c.

Following Gao \etal~\cite{gao2020structured}, we also include modality pair labels [\textit{is\_obj\_to\_obj}, \textit{is\_obj\_to\_ocr}, \textit{is\_ocr\_to\_obj}, \textit{is\_ocr\_to\_ocr}] that indicate the modalities of each pair of objects. For example, the \textit{is\_obj\_to\_ocr} label is positive if the `self' object is a visual object recognized by the Faster R-CNN and the `other' object is an OCR token. These labels will help the model learn different interactions between modalities. This edge feature is illustrated in Figure \ref{fig:edge_features}d.

Finally, these features are only meaningful for pairs of Transformer inputs that originate from the image (the visual and OCR modalities). Therefore, when a question modality Transformer input is involved, we set the aforementioned features to be zero.

\vspace{-2mm} 
\paragraph{Vanilla Transformer Self-Attention}
For some object or token $i$, the Transformer \cite{attention_is_all_you_need} self-attention module computes the attended representation of each of $n_{obj}$ objects as

\begin{equation}
    \text{Attention}(\textbf{q}_i, K, V) = \text{softmax}\left(\frac{\textbf{q}_i K^T}{\sqrt{d_k}}\right)V
\end{equation}

where $\textbf{q}_i$, a row vector of dimension $d_k$, is the query for object $i$. $K$ is a matrix composed of $n_{obj}$ stacked row vectors of dimension $d_k$, representing the keys of the $n_{obj}$ objects. $V$ is a matrix composed of $n_{obj}$ stacked row vectors of dimension $d_v$, representing the values of the $n_{obj}$ objects.

\begin{figure*}[!h]
     \centering
    \includegraphics[width=\linewidth]{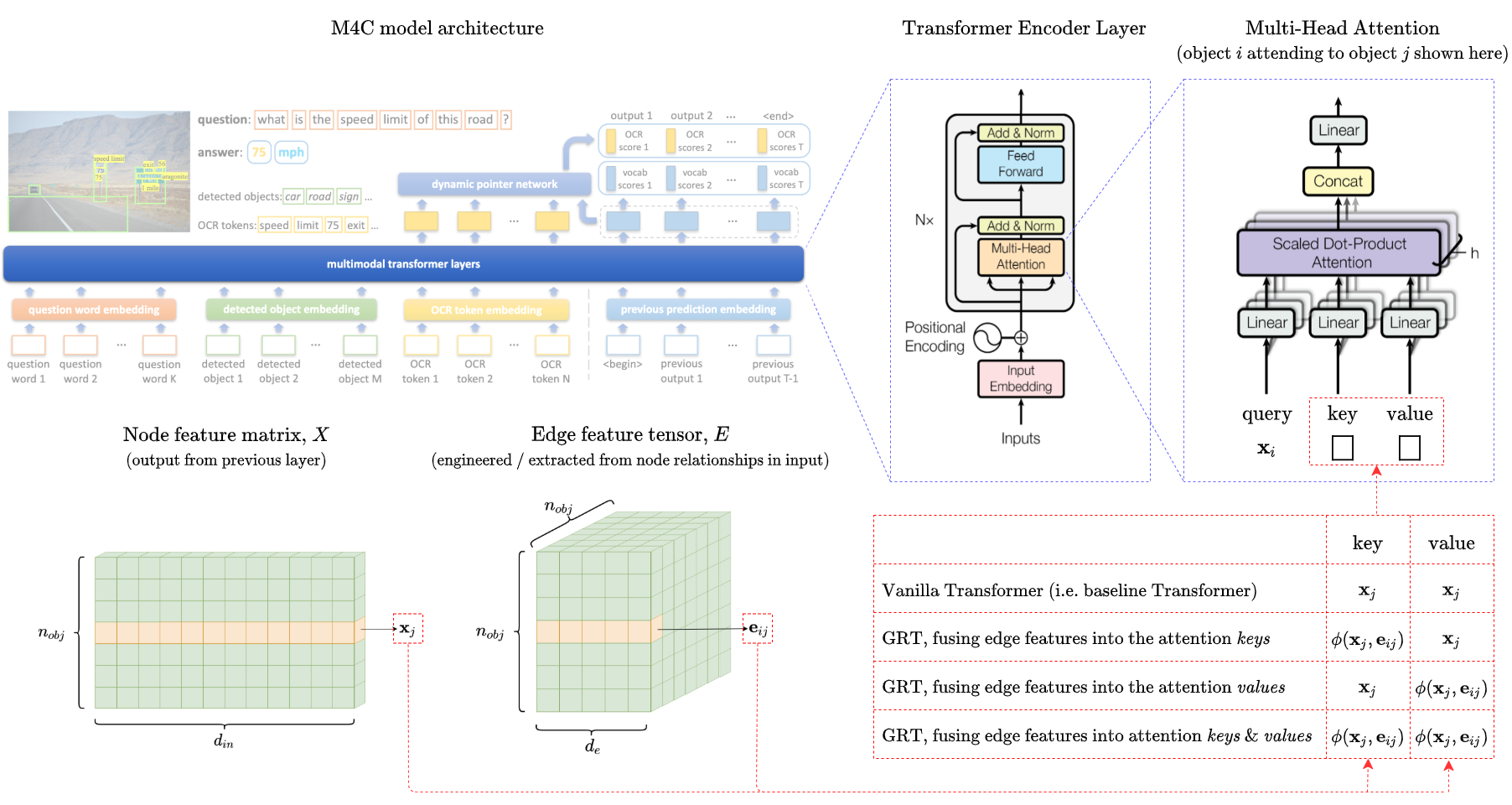}
    
    \caption{Illustration of how the Graph Relation Transformer (GRT) extends the `Vanilla' Transformer Architecture \cite{attention_is_all_you_need}, noting where this module can be inserted into the M4C model for the TextVQA task. We propose that edge features $\textbf{e}_{ij}$ between objects $i$ and $j$ can be incorporated using a fusion function $\phi$ into the keys and/or values of the Transformer Multi-Head Attention module. $\phi_{concat}$ (concatenation) or $\phi_{add}$ (addition) are possible fusion functions that can be used, for example. Elements of this illustration are adapted from figures in \cite{m4c} and \cite{attention_is_all_you_need}.}\label{fig:grt_diagram}
\end{figure*}

The node features matrix $X$ comprises $n_{obj}$ row vectors of node features of dimension $d_{in}$ (visualized in Figure \ref{fig:grt_diagram}).\footnote{In the first Transformer layer, $X$ is simply the input features, whereas in subsequent layers, $X$ is the output of the previous layer.} We obtain $K$ and $V$ using projection matrices $W_k$ and $W_v$ to project the $n_{obj}$ rows of $X$ to vector spaces of dimension $d_k$ and $d_v$ respectively.
\begin{equation}\label{eqn:K}
    K = XW_k
\end{equation}
\begin{equation}\label{eqn:V}
    V = XW_v
\end{equation}
Notice how there is no place to incorporate edge features and that the same $K$ and $V$ matrices are used to compute $\text{Attention}(\textbf{q}_i, K, V)~\forall i$.

\paragraph{Graph Relation Transformer Self-Attention}
We would like to fuse the $n_{obj}^2$ pairwise edge features between each of the $n_{obj}$ objects into the self-attention module. Suppose we have our edge feature tensor $E$ of dimension $n_{obj} \times n_{obj} \times d_e$, where the vector $E_{ij}$, which we will henceforth refer to as $\textbf{e}_{ij}$, represents the $d_e$-dimensional edge feature between objects $i$ and $j$ with $i$ as the `self' object and $j$ as the `other' object (the edges are directional). Figure \ref{fig:grt_diagram} visualizes this $E$ tensor. Essentially, $E$ is a stack of $n_{obj}$ matrices, where each of these matrices, $E_i$, represents the $n_{obj}$ sets of edge features where object $i$ is the `self' object. These features may be any fixed number of dimensions of discrete and/or continuous variables.

We propose three different places to fuse these features in the Transformer architecture: (1) in the keys, (2) the values and (3) both the keys and values. We illustrate this fusion in the table present in Figure \ref{fig:grt_diagram}. Two central ideas motivate fusion in the keys and the values. On the one hand, fusing in the keys makes object $i$'s attention weight for object $j$ depend on $\textbf{e}_{ij}$. This means that the \textit{relevance} of object $j$ on the representation of object $i$ should depend on how the two objects are related. On the other hand, because self-attention ultimately computes a weighted average of the values, fusing in the values makes the \textit{influence} of object $j$ on the attended representation of object $i$ depend on $\textbf{e}_{ij}$

We will now define this formally. When we fuse the edge features in the keys, we now compute a different set of keys, $K_i$, for each object $i$ in order to allow the relations between each object $i$ with all other objects to vary for different $i$. This is defined in the Equation \ref{eqn:fuse_keys}.
Similarly, fusing the edge features in the values via $V_i$ is defined in Equation \ref{eqn:fuse_values}. Finally, Equation \ref{eqn:fuse_keysvalues} defines fusing in both keys and values.
\begin{equation}
    \label{eqn:fuse_keys}
    \text{GraphAttn}_{k}(\textbf{q}_i, K, V) = \text{softmax}\left(\frac{\textbf{q}_i K_i^T}{\sqrt{d_k}}\right)V
\end{equation}
\begin{equation}
    \label{eqn:fuse_values}
    \text{GraphAttn}_{v}(\textbf{q}_i, K, V) = \text{softmax}\left(\frac{\textbf{q}_i K^T}{\sqrt{d_k}}\right)V_i
\end{equation}
\begin{equation}
    \label{eqn:fuse_keysvalues}
    \text{GraphAttn}_{kv}(\textbf{q}_i, K, V) = \text{softmax}\left(\frac{\textbf{q}_i K_i^T}{\sqrt{d_k}}\right)V_i
\end{equation}
Now, in order to obtain the fused keys $K_i$, we fuse the object features $X$ with $E_i$ using a fusion function $\phi$ before projecting the representation to the `keys' vector space.
\begin{equation}\label{eqn:K_i}
    K_i = \phi(X, E_i)W_k
\end{equation}
In the same way, we can fuse object features with edge features in the values by 
\begin{equation}\label{eqn:V_i}
    V_i = \phi(X, E_i)W_v
\end{equation}
Notice the difference between Equations \ref{eqn:K}-\ref{eqn:V} and \ref{eqn:K_i}-\ref{eqn:V_i}. In this work, we propose two fusion functions: $\phi_{concat}$ and $\phi_{add}$.
\begin{equation}\label{eqn:phi_concat}
    \phi_{concat}(X, E_i) = [X;E_iW_c]
\end{equation}
is the concatenation fusion function where $[;]$ is the concatenation operation and the $d_e \times d_{e'}$ matrix $W_c$ projects $E_i$ to an intermediate vector space of dimension $d_{e'}$. We include this projection to give each Transformer layer the flexibility to use the edge features differently. However, one could set $W_c$ to the identity matrix to effectively disable this projection.\footnote{Clearly, if $W_c$ or $W_a$ are not the identity matrix, then the fusion function itself will also contain parameters / weights to be learned along with the rest of the model.} And
\begin{equation}\label{eqn:phi_add}
    \phi_{add}(X, E_i) = X + E_i W_a
\end{equation}
is the addition fusion function where the $d_e \times d_{in}$ matrix $W_a$ projects $E_i$ to the same space as $X$ so that addition can be performed.

\subsection{Parameter Learning}\label{subsec:parameter_learning}
The GRT is an extension to the baseline M4C model and the extra parameters we introduce are trained end-to-end along with the existing parameters using the same multi-token sigmoid loss as M4C. We define this loss formally in Equation \ref{eqn:bce}, where $N$ is the number of training examples, $M_i$ is the number of tokens in the ground truth answer for example $i$, $y_{ij}$ is the vocabulary index of the $j^{\text{th}}$ answer token of the $i^{\text{th}}$ example and $\hat{y}_{ij}$ is the the model's sigmoid activation value for that token.
\begin{equation}\label{eqn:bce}
    \mathcal{L} = -\frac{1}{N}\sum_{i=1}^{N} \frac{1}{M_i} \sum_{j=1}^{M_i} ({y}_{ij}  \log \left(\hat{y}_{ij}\right) + (1-{y}_{ij}) \log \left(1-\hat{y}_{ij}\right))
\end{equation}

The extra parameters introduced by GRT are those used by $\phi$ (i.e. $W_c$ or $W_a$) inside the Transformer layers.

\section{Experimental Setup}
\label{experimental_setup}

\subsection{Dataset} We use the TextVQA dataset \cite{lorra} for this task since the TextVQA dataset is the official dataset for the TextVQA task. The TextVQA dataset contains 45,336 questions on 28,408 images where all questions require reading and reasoning about the text in images. The dataset contains images gathered from the Open Images dataset \cite{krasin2017openimages} and contains images which contain text like traffic signs, billboards etc. We use the standard TextVQA dataset split for dividing the dataset into training, validation and testing sets. The training set consists of 34,602 questions from 21,953 images, the validation set consists of 5000 questions from 3166 images and the test set consists of 5734 questions from 3289 images. The TextVQA dataset contains visual (images) and text (questions) modalities. In addition to this, the dataset also provides Rosetta \cite{borisyuk2018rosetta} OCR tokens which forms the additional OCR modality in the task. Each question-image pair has 10 ground-truth answers annotated by humans. 

\subsection{Evaluation Metric} We use the evaluation metric given by the VQA v2.0 challenge \cite{goyal2017making}. This metric does some preprocessing before looking for exact matches between the human annotated answers and the model’s output. This metric averages all of the 10 choose 9 sets of the human annotated answers as they determined this approach is more robust to inter-human variability in phrasing answers. The evaluation metric equation is shown below where `a' is the model’s output:
\begin{equation}
\operatorname{Acc}(a)=\min \left\{\frac{\# \text { humans \, that  \, said \,  \textit{a}}}{3}, 1\right\}
\end{equation}
\subsection{Baseline model: M4C}
We build upon the M4C baseline model for all our experiments. The M4C model first projects the feature representations of the different entities (question words, objects, OCR tokens) into a common embedding space. The question words are embedded using a pre-trained BERT model. The visual objects are represented using Faster R-CNN features (appearance ($\textbf{x}_{i}^{\mathrm{fr}}$) and location features ($\textbf{x}_{i}^{\mathrm{b}}$) for some object $i$) which are then projected onto a common embedding space, followed by layer normalization (LN) and addition as shown in Equation \ref{eq1}:
\begin{equation}
\label{eq1}
\textbf{x}_{i}^{\mathrm{obj}}=\mathrm{LN}\left(W_{1} \textbf{x}_{i}^{\mathrm{fr}}\right)+\mathrm{LN}\left(W_{2} \textbf{x}_{i}^{\mathrm{b}}\right)
\end{equation}
M4C uses the Rosetta OCR system to recognize OCR tokens in the image along with their location. For some recognized OCR token $j$, a 300-dim FastText vector ($\textbf{x}_{j}^{\mathrm{ft}}$), appearance feature from Faster-RCNN ($\textbf{x}_{j}^{\mathrm{fr}}$), a 604-dim Pyramidal Histogram of Characters (PHOC) vector ($\textbf{x}_{j}^{\mathrm{p}}$) and a 4-dim location vector ($\textbf{x}_{j}^{\mathrm{b}}$) are extracted. Similar to the visual object features, these are projected to the common embedding space to get the final OCR feature for each token as shown in Equation \ref{eq2}:
\begin{equation}
\label{eq2}
\textbf{x}_{j}^{\mathrm{ocr}}=\mathrm{LN}\left(W_{3} \textbf{x}_{j}^{\mathrm{ft}}+W_{4} \textbf{x}_{j}^{\mathrm{fr}}+W_{5} \textbf{x}_{j}^{\mathrm{p}}\right)+\mathrm{LN}\left(W_{6} \textbf{x}_{j}^{\mathrm{b}}\right)
\end{equation}

After embedding all the modalities, they are all passed through a multi-layer Transformer with multi-headed self-attention. Each entity here can attend to all entities in the same modality and in other modalities. The output from the Transformer is then passed through a pointer network to iteratively decode the answer. At each time step, the pointer network chooses a token from either the fixed training vocabulary or the OCR tokens by taking in the embedding of the previously predicted word. The model is trained using a multi-label sigmoid loss as shown in Equation \ref{eqn:bce}.  

\subsection{Experimental Methodology for the GRT}
  
Since the GRT builds upon the existing M4C baseline, we used the same initialization method as M4C for each of the parameters within our model. Due to the limited access of the TextVQA test set evaluation server, we used the TextVQA evaluation metric to compare each of the methods accuracy on the validation set. To evaluate and compare the best graph attention fusion methods and fusion locations, we trained a model for each scenario for 5,000 update steps to determine which of the graph attention setups was the best before training each model fully to convergence. Once the best graph attention setup was chosen, this became our Graph Relation Tranformer shown in Table \ref{tab:results_summary}. We trained the GRT method for 24,000 updates for a fair comparison as this was the number of updates originally used by the M4C authors \cite{m4c}. Additionally, we conducted an ablation study in which we removed one of the edge feature types and trained the model until convergence to provide clarity to the information content gained by each of the edge feature types.

All our models use 4 Transformer layers with 12 attention heads.\footnote{Note that in previous work \cite{gao2020structured,kant2020spatially}, simply increasing the number of Transformer layers has been found to improve performance. We have not reproduced this tweak as it is not relevant to the ideas we explore in this paper.} 
Answer decoding is done for a maximum of 12 tokens. 
On four V100 GPUs, training the baseline M4C model for 24,000 updates took 5 hours. 
Adding the computation necessary to include all our edge features increased this training time to 8 hours when fusing them in either the keys or values and 12 hours in both. 
However, the appearance embedding cosine similarity was a computationally expensive feature and if it is removed, the time is reduced to 6 hours when fusing edge features in values only.



\section{Results and Discussion}
\label{results}

\begin{table}[hbt!]
    \centering
    \begin{tabular}{lcccc}
        \toprule
        Architecture & Opts & Val (\%) & $\Delta$Val (\%) & Test (\%)\\
        \midrule
        LoRRA$^*$ &  & 27.17 & - & \\
        M4C$^*$ &  & 38.93 & 0 & N/A\\
        GRT$^*$ &  & 39.58 & \textbf{0.65}& 39.58\\
        \midrule
        M4C &  & 39.40 & 0 & 39.01\\
        SMA &  & 39.58 & \textbf{0.18} & 40.29\\
        \midrule
        M4C & \cmark & 42.7 & 0 & N/A\\
        SAMT & \cmark & 43.90 & \textbf{1.20} & N/A\\
        \bottomrule
    \end{tabular}
    \caption{\label{tab:results_summary} Performance comparison on the TextVQA validation('Val') and Test('Test') sets. $\Delta$Val refers to the improvement of a model's accuracy over the M4C model trained with no optimizations other than the architectural change proposed by the model. Rows marked with * are results we obtained ourselves by using the \texttt{MMF} \cite{singh2020mmf} framework with default settings. The raw SAMT accuracy values are due to both their novel architecture as well as other optimizations (Opts), namely adding 2 extra Transformer layers, using Google OCR as the OCR module and a ResNext-152 \cite{resnext152} Faster R-CNN model \cite{faster_rcnn} trained on Visual Genome \cite{visual_genome}. The GRT model shown here uses the `spatial translation', `spatial interaction labels' and `modality pair labels' edge features and it fuses these edge features in the values of the Multiheaded Attention Module by addition (the $\phi_{add}$ function).
    }
\end{table}

As seen in Table \ref{tab:results_summary}, the Graph Relation Transformer architecture surpasses the LoRRA and M4C baseline models. We also show results for the SMA \cite{gao2020structured} and SAMT \cite{kant2020spatially} models as originally reported by their authors.

Compared to the experimental settings originally used by the M4C authors Hu \etal~\cite{m4c}, the authors of SMA and SAMT report results on models that include not only the novel architectural changes they propose, but also other optimizations. Namely, SMA uses a custom Rosetta-en OCR module and SAMT uses 2 extra Transformer layers, a Google OCR module and a ResNext R-CNN. Therefore, we show the improvement in accuracy over the corresponding M4C model trained with the same optimizations. SMA and SAMT achieve 0.18\% and 1.2\% accuracy increases over their corresponding M4C models respectively. 

From our experiments, modifying the baseline M4C model with the GRT architecture alone yields a 0.65\% performance increase. This result is from the best model setup determined from the Transformer fusion exploration and the edge feature ablation study. This best model fused the edge features into the values location, used the $\phi_{add}$ fusion function and used all edge feature types except for the visual similarity edge feature type.  

The same model (coincidentally) achieved 39.58\% accuracy on both the test set (evaluated by the official test server) and validation set, submitted as team \texttt{cmu\_mmml}.


\subsection{Fusion Function for Edge Features} 

\begin{table}[hbt!]
    \centering
    \begin{tabular}{llc}
        \toprule
        Fusion Location & Fusion function $\phi$ & Accuracy (\%)  \\
        \midrule
        Keys & $\phi_{concat}$ & 35.51 \\
        Keys & $\phi_{add}$ & 35.17 \\
        Values & $\phi_{concat}$ & 36.90 \\
        Values & $\phi_{add}$ & \textbf{37.30} \\
        Keys \& Values & $\phi_{concat}$ & 31.89 \\
        Keys \& Values & $\phi_{add}$ & 35.41 \\
        \bottomrule
    \end{tabular}
    \caption{\label{tab:model_selection} Performance on the TextVQA validation set of Graph Relation Transformer with different methods of fusing the edge features. As the purpose of these runs was model selection only, training for each model was stopped after 5,000 steps (before convergence). The fusion functions are defined in Equations \ref{eqn:phi_concat}-\ref{eqn:phi_add}.}
\end{table}

\begin{figure*}[hbt!]
    \centering
    \subfloat[
    What word is printed under interior design on the  book in the middle?\\
    M4C: para\\
    Graph Relational Transformer: inspirations\\
    Best answer: inspirations
 ]{{\includegraphics[height=3.8cm,width=0.32\textwidth]{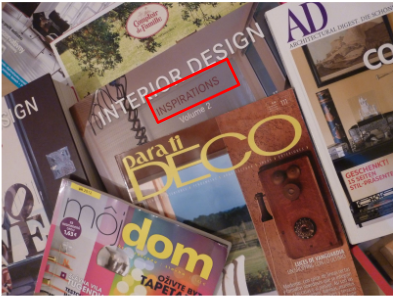}}}
    \hspace{2mm}
    \subfloat[
    What kind of establishment is in the background next to the red and white truck?\\
    M4C: plus \\
    Graph Relational Transformer: bar\\
    Best answer: bar
    ]{{\includegraphics[height=3.8cm,width=0.34\textwidth]{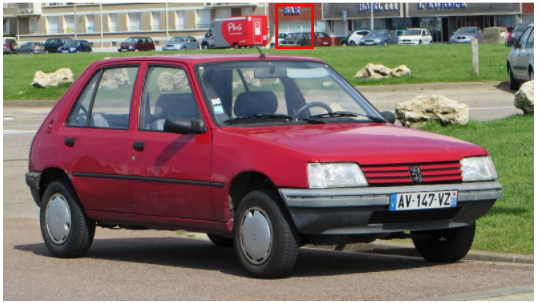}}}
    \hspace{2mm}
    \subfloat[
    What company is on the left side of the screen?\\
    M4C: comerica\\
    Graph Relational Transformer: meijer\\
    Best answer: meijer
    ]{{\includegraphics[height=3.8cm,width=0.28\textwidth]{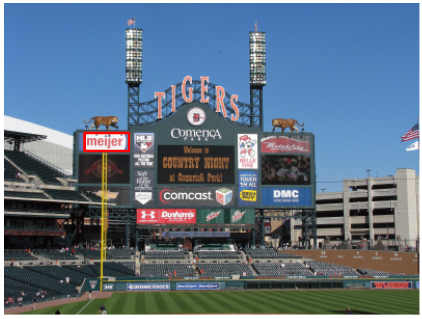}}}
    
    \caption{Qualitative Analysis: The Figure shows qualitative examples of illustrative cases comparing the M4C model and the Graph Relational Transformer. The red boxes have been applied for this figure specifically and are not present in the original image. `Best answer' refers to the consensus human answer in the dataset.}\label{fig:graph_model_qual}
\end{figure*}

To determine the best fusion location and fusion function, we compared all of the possible combinations shown above in Table \ref{tab:model_selection}. The intuition behind the keys only fusion location is that the importance given to another object would be affected by the spatial relationships between the objects. The intuition behind the values only fusion location is that the representation of an object would change depending on the spatial relationship between the objects. As shown above, the values only fusion location and the $\phi_{add}$ fusion function performed the best. This shows that the model was better able to reason about the image when the model was able to change the representation of another object based on the context and spatial relationship between the current object.

\subsection{Ablation Study on Edge features} 
\label{ablation}

\begin{table}[hbt!]
    \centering
    \begin{tabular}{lc}
        \toprule
        Feature set & Accuracy (\%)  \\
        \midrule
        All features & 39.40 \\ 
        - Appearance similarity & \textbf{39.58} \\
        - Spatial translation feature & 39.06 \\
        - Spatial interaction labels & 38.78 \\
        - Modality pair labels & 38.96 \\
        \bottomrule
    \end{tabular}
    \caption{\label{tab:ablation_study} Ablation Study on Edge Features: Performance on the TextVQA validation set of Graph Relation Transformer with various groups of features ablated. Each model fuses edge features at self-attention `values' only, using `add' as the fusion function $\phi$. Each model was trained 24,000 steps (to convergence).}
\end{table}
To evaluate the impact of each of the edge feature types, an ablation study was done for each edge feature type as shown in Table \ref{tab:ablation_study}. We observe that while performance suffered when most of our edge features were dropped (as expected), the one exception was that when the appearance similarity edge feature was dropped, performance actually improved. Recall that our appearance similarity feature is cosine similarity, which is nothing more than a (scaled) dot product between the visual embeddings from the R-CNN. Our explanation for why this cosine similarity does not improve performance is that in the baseline Transformer architecture, the attention module already does a scaled dot product between (projections of) the visual embeddings. Therefore, the cosine similarity does not add anything meaningful to the model apart from introducing noise.

\subsection{Qualitative Analysis}

We further performed a qualitative comparison between the predictions of the M4C and the Graph Relational Transformer to identify instances that  benefited from injecting object relational features into the transformer. Figure \ref{fig:graph_model_qual} shows some typical examples which demonstrate the effectiveness of the Graph Relational Transformer. Figure \ref{fig:graph_model_qual}a and Figure \ref{fig:graph_model_qual}b provides instances where the Graph Relational Transformer was better able to reason about the relation between the OCR and objects in the image to generate the correct answer. Figure \ref{fig:graph_model_qual}c provides one instance where the Graph Relational Transformer was able to answer questions involving positions better that the M4C baseline. Here, the spatial translation feature and the modality pair labels give useful information to the model to answer these kinds of questions as they allow the model to build a superior representation of the tokens by selectively interacting with the other tokens based on their relationships.

\section{Conclusion and Future Directions} 
\label{conclusion}


In this work we propose a Graph Relation Transformer for TextVQA. Our best performing model uses the $\phi_{add}$ fusion function at the attention `values', and incorporates all of the edge features except for the visual similarity edge feature. It outperforms the M4C model due to its improved spatial reasoning ability. Our quantitative and qualitative results support our hypothesis that incorporating spatial relationships between objects in the image lead to better performance. Additionally, most of the errors in the baseline and our models were due to the OCR system incorrectly detecting words within the image. Hence, we believe that there is significant performance headroom from improving the OCR system, as accurate OCR tokens are the first crucial step in reasoning about the image. Beyond TextVQA, we have noted that the GRT can be applied to any task where relations between objects can be represented by vectors. Exploring the TextVQA dataset further, possible future work includes exploring additional edge feature types and using an improved OCR system with the GRT architecture. Beyond these ideas, applying the GRT architecture to different datasets and tasks present additional potential research directions. We also note that the GRT architecture is generalizable to any application or task with one or more modalities where relations between objects are informative. We leave the validation of this hypothesis to future work.

{\small
\bibliographystyle{ieee_fullname}
\bibliography{egbib}
}

\end{document}